\documentclass[10pt,twocolumn,letterpaper]{article}

\usepackage{cvpr}
\usepackage{times}
\usepackage{epsfig}
\usepackage{graphicx}
\usepackage{amsmath}
\usepackage{amssymb}

\usepackage[pagebackref=true,breaklinks=true,letterpaper=true,colorlinks,bookmarks=false]{hyperref}
\cvprfinalcopy 

\ifcvprfinal\fi
\begin{document}

\title{Facial Image Deformation Based on Landmark Detection}

\author{Chaoyue Song, Yugang Chen, Shulai Zhang and Bingbing Ni\\
Shanghai Jiao Tong University, China\\
{\tt\small \{beyondsong, cygashjd, zslzsl1998, nibingbing\}@sjtu.edu.cn}
}
\maketitle

\begin{abstract}
In this work, we use facial landmarks to make the deformation for facial images more authentic. The deformation includes the expansion of eyes and the shrinking of noses, mouths, and cheeks. An advanced 106-point facial landmark detector is utilized to provide control points for deformation. Bilinear interpolation is used in the expansion and Moving Least Squares methods (MLS) including Affine Deformation, Similarity Deformation and Rigid Deformation are used in the shrinking. We compare the running time as well as the quality of deformed images using different MLS methods. The experimental results show that the Rigid Deformation which can keep other parts of the images unchanged performs better even if it takes the longest time.
\end{abstract}

\section{Introduction}
Image deformation, as one of the most popular topics in the area of computer vision and image processing, has been discussed for many years. Recently, the emergence of artificial intelligence has enabled new techniques in image deformation and has achieved impressive achievements, especially in some specific scenarios.  The deformation aimed for faces is one of the most popular areas in academe as well as industry. With more and more people pursuing beauty, verisimilar deformed facial images and an automatic process to generate these images are required to meet these people's needs. Accurate deformation methods are continuously in great demand. Before the emergence of artificial intelligence, common manipulations (e.g. expansion, shrinking, and blurring) on facial features have no differences from that on other objects. This is because those manipulations omit special characteristics of facial features. Deep learning methods can extract facial landmarks from facial images, which provides possibilities of warping based on these landmarks. With these landmarks, we can produce more accurate results and make the warped image more authentic. There are amounts of techniques in operating deformations onto facial images and the fundamental operation is image warping, especially
\begin{figure}[htbp]
\begin{center}

   \includegraphics[width=1.0\linewidth]{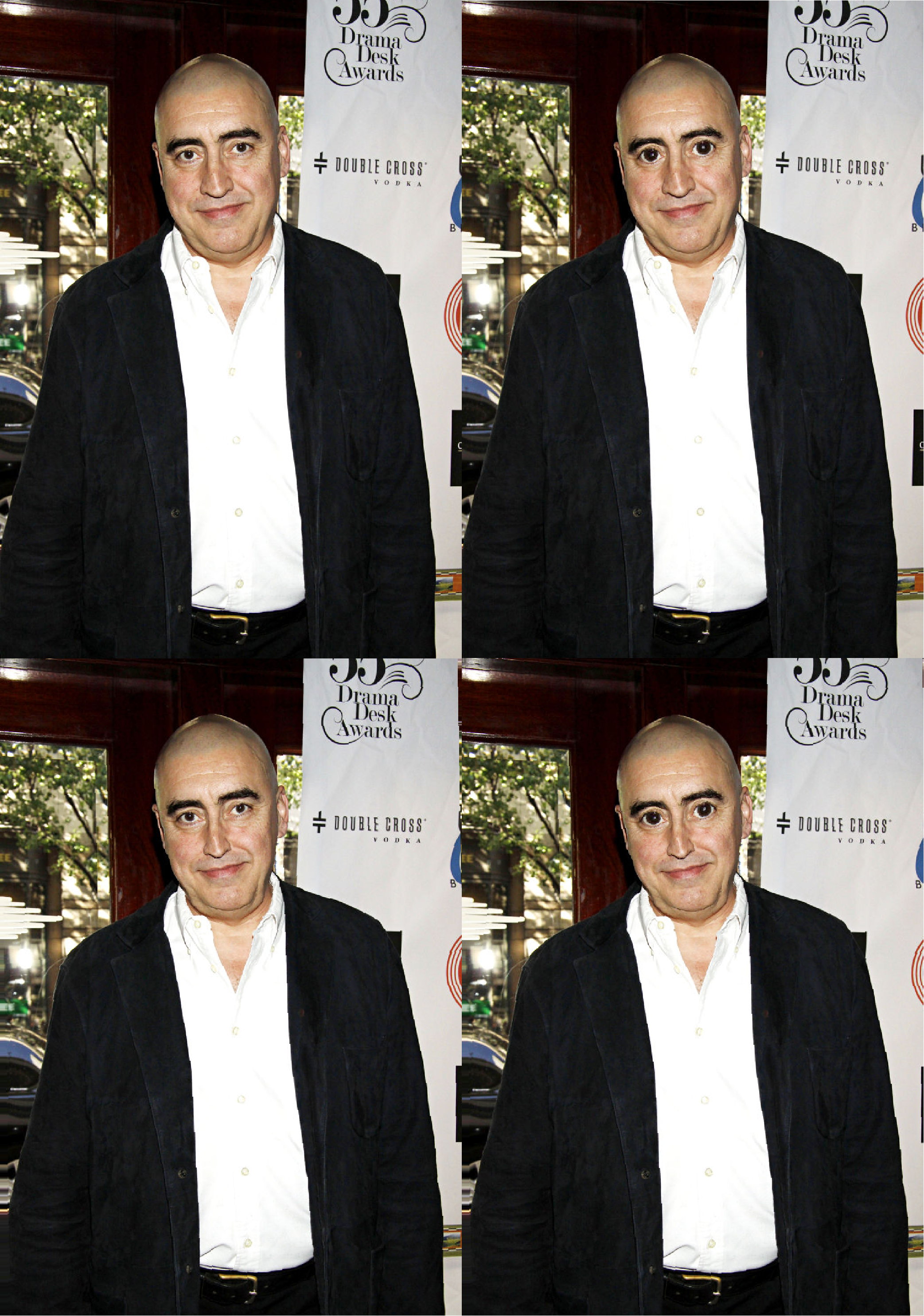}
\end{center}
   \caption{Examples of our deformation result. From top left to bottom right: original image, image with eye expansion, image with nose, mouth, and cheek shrinking, image with both expansion and shrinking.}
\label{ourresult}
\end{figure}
warping with control points.

Expansion and shrinking are two preferred operations on facial images. Expansion is often used in deformations onto eyes and shrinking is often used on noses and mouths. For an automatic facial deformation process, it is intuitive that more landmarks are detected accurately and more authentic deformed images will be obtained. Dlib\footnote{http://dlib.net/} can provide a 68-point facial landmark detection. Nevertheless, that is not enough. For example, there are too few points located near the nose which makes it difficult to perform deformations.
Thus, to perform deformations with more accuracy, we need a landmark detector that provides more control points.

The key in the deformation stage is to find an accurate mapping function to map one reference point or several adjacent points if required to the wanted point, based on several control points. In addition, we may need to choose different mapping functions according to the demands of different deformations.

In this work, we trained an advanced 106-point facial landmarks detector based on the method proposed by \cite{zhou2013extensive}, which can provide enough control points for the deformation. Then we implemented the expansion based on bilinear interpolation whose degree of expansion is adjustable. Shrinking is achieved by the Moving Least Squares algorithm (MLS) \cite{schaefer2006image} which includes Affine Deformation, Similarity Deformation and Rigid Deformation. The experimental result shows that our method which combines facial landmark detection and image deformation can provide authentic deformed facial images.

\section{Related Work}

\paragraph{Facial Landmark detection}
The research on facial landmark detection can trace back to 1995 when an Active Shape Model (ASM) \cite{cootes1995active} was proposed. ASM is based on Point Distribution Model and it was improved into Active Appearance Models \cite{cootes1998active} which consists of Shape Model and Texture Model. With the development of deep learning, convolutional neural network is used in facial detection for the first time \cite{sun2013deep}. In \cite{sun2013deep}, a deep convolutional neural network (DCNN) is proposed. Later, Face++ \cite{zhou2013extensive} improved the accuracy in DCNN and can detect and localize more landmarks with higher accuracy. After that, TCDCN (Tasks-Constrained Deep Convolutional Network) \cite{zhang2014facial}, MTCNN (Multi-task Cascaded Convolutional Networks) \cite{zhang2016joint}, TCNN(Tweaked Convolutional Neural Networks) \cite{wu2018facial}, DAN(Deep Alignment Networks) \cite{kowalski2017deep} came up and performed better and better. Recently, the work \cite{merget2018robust}, \cite{song2018beyond}, and \cite{wu2018look} have proposed more methods with higher accuracy and robustness.
\vspace{-10pt}
\paragraph{Image Warping}
Image warping is a transformation which maps all positions in one image plane to positions in a second plane \cite{glasbey1998review}. There are three forms of warping which are translation warping, scaling warping and rotation warping in \cite{wolberg1990digital}. The common ground for image warping is that a set of handles (also named as control points) is required. However, these methods do not consider the features of the image. In \cite{beier1992feature}, a feature-based image deformation method is proposed to solve these problems. And an image deformation method based on linear Moving Least Squares was proposed in \cite{schaefer2006image} to meet the smoothness, interpolation and identity demands for image deformation. Such deformation has the property that the amount of local scaling and shearing is minimized. Later, an image warping method based on artificial intelligence was proposed in \cite{shiraishi2011computer} and techniques in artificial intelligence are used more widely in image warping.

\section{Method}
\subsection{Facial Landmark Extraction}
\begin{figure}[t]
\begin{center}

   \includegraphics[width=0.8\linewidth]{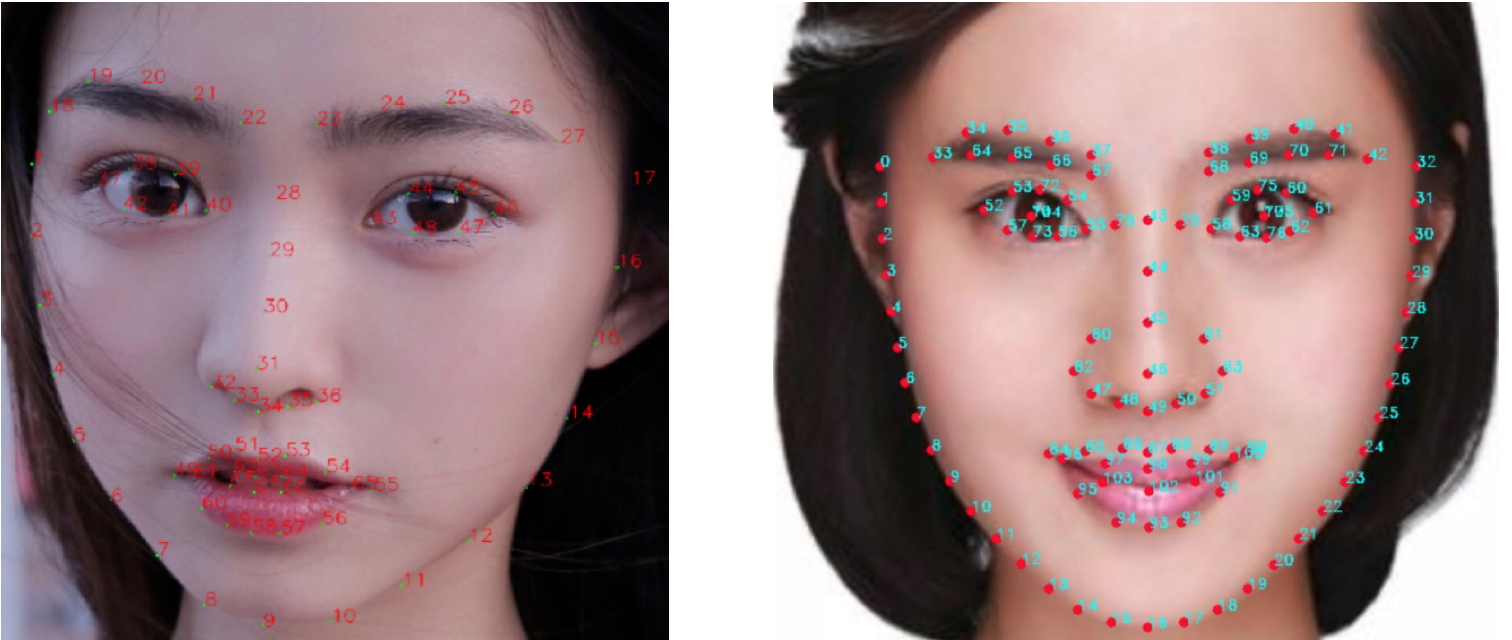}
\end{center}
   \caption{Facial landmark detection. Left: 68-point facial landmark detection. Right: 106-point facial landmark detection.}
\label{landmark}
\end{figure}

Accurate facial landmark extraction is the prerequisite of successful facial image deformation. The model we use in this work is proposed in \cite{zhou2013extensive}. We improve the 68-point facial landmark detector provided by Dlib to a 106-point one. As shown in Figure \ref{landmark}, there are 33 landmarks for cheeks, 18 landmarks for eyes, 18 landmarks for eyebrows, 15 landmarks for the nose, and 20 landmarks for the mouth.

\subsection{Expansion}
\begin{figure}[t]
\begin{center}

   \includegraphics[width=1\linewidth]{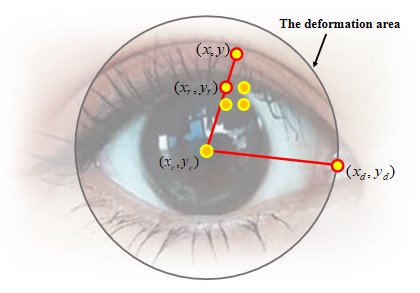}
\end{center}
   \caption{Scheme one for eye expansion. Using the center landmark as $E_c$. The four orange points whose outlines are yellow are the four pixels used in bilinear interpolation. The intervals between pixels are exaggerated.}
\label{expansion1}
\end{figure}

\begin{figure}[t]
\begin{center}

   \includegraphics[width=1\linewidth]{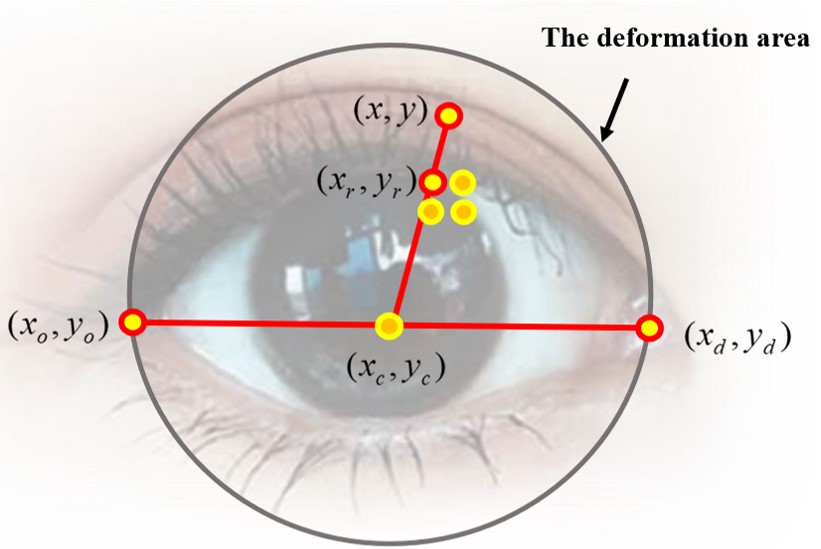}
\end{center}
   \caption{Scheme two for eye expansion. Using the middle point between $E_o$ and $E_d$ as $E_c$. The four orange points whose outlines are yellow are the four pixels used in bilinear interpolation.}
\label{expansion2}
\end{figure}

In this section, we take the manipulation on the left eye $E$ as an example for explanation. The manipulation on the right eye is exactly symmetric to the manipulation on the left eye. Before doing expansion, we need to first determine the control points, thereby making sure the area that needs to be adjusted, which is named as the deformation area in this paper. The center point $E_c(x_c,y_c)$ and the landmark $E_d(x_d,y_d)$ at the corner of eye $E$ are used to determine the boundary of the deformation area. There are two schemes to determine $E_c$. One is regarding the center landmark as $E_c$, and the other is regarding the midpoint between the landmark for the outer canthus $E_o$ and the landmark for the inner canthus $E_d$ as $E_c$. These two schemes are illustrated in Figure \ref{expansion1} and \ref{expansion2}.

The deformation area is a circle, whose center is $E_c$ and radius $R_E=((x_d-x_c)^2+(y_d-y_c)^2)^{1/2}$. Thus, the pixel $P(x,y)$ within the deformation area satisfies $((x-x_c)^2+(y-y_c)^2)^{1/2}<R_E$. For each pixel in the deformation area, there is a corresponding reference pixel $P_r(x_r,y_r)$. To find the corresponding reference pixel, we define a parameter $a$ which is used to determine the expansion scale for each pixel $S(x,y)$. Then the expansion scale is expressed as

\begin{equation}
S(x,y)=1-\frac{a}{100}\times(1-\frac{(x-x_c)^2+(y-y_c)^2}{R_E^2})    
\end{equation}

The reference pixel's position is 
\begin{equation}
 x_r=(x-x_c)\cdot S(x,y)+x_c   
\end{equation}

\begin{equation}
y_r=(y-y_c)\cdot S(x,y)+y_c    
\end{equation}

We use bilinear interpolation to compute the value of $P(x,y)$ after deformation, which is expressed as $f(x,y)$. Then

\begin{equation}
    f(x,y)=\frac{\sum\limits_{i}\sum\limits_{j}f(x_r+i,y_r+j)(x-x_r-i)(y-y_r-j)}{\sum\limits_{i}\sum\limits_{j}(x-x_r-i)(y-y_r-j)}
\end{equation}
where $i$ and $j$'s values are $0$ and $1$.

\subsection{Shrinking}

The methods we use in the shrinking are Moving Least Squares (MLS) \cite{schaefer2006image}. 

\subsubsection{Background}
\hspace{1.0em}
MLS views the deformation as a function $f$ that maps all pixels in the undeformed image to pixels in the deformed image. It needs to apply the function $f$ to each point $v$ in the undeformed image. In \cite{schaefer2006image}, the authors consider building image deformations based on collections of points with which the user controls the deformation and this is natural in facial images. Let $p$ be a set of control points and $q$ be the deformed positions of the control points. 

For a point $v$ in the image, the best affine transformation $l_v(x)$ can be solved by minimizing

\begin{equation}
\sum_i w_i|l_v(p_i)-q_i|^2    
\end{equation}
where $p_i$ and $q_i$ are row vectors and the weights $w_i$ have the form

\begin{equation}
w_i=\frac{1}{|p_i-v|^{2\alpha}}    
\end{equation}

Therefore, a different transformation $l_v(x)$
for each $v$ can be obtained.

Next, the deformation function $f$ can be defined to be $f(v) = l_v(v)$. There are three kinds of functions that will induct different deformation effects which are Affine Deformation, Similarity Deformation and Rigid Deformation. The mapping function for Affine Deformation is 

\begin{equation}
f_a(v) = (v-p_*)(\sum_i \hat{p}_i^T w_i \hat{p}_i)^{-1} \sum_j \hat{p}_j^T 
\hat{q}_j+q_*    
\end{equation}

The mapping function for the Similarity Deformation is

\begin{equation}
f_s(v)=\sum_i \hat{q}_i (\frac{1}{\mu_s}A_i) + q_*    
\end{equation}
where $\mu_s = \sum_i w_i \hat{p}_i \hat{p}_i^T $ and $A_i$ depends only on the $p_i$, $v$ and $w_i$ which can be precomputed. $A_i$ is

\begin{equation}   
A_i=w_i
\left(                 
  \begin{array}{c}

    \hat{p}_i\\
    -\hat{p}_i^{\bot}\\

  \end{array}
\right)
\left(                 
  \begin{array}{c}

    v-p_*\\
    -(v-p_*)^{\bot}\\

  \end{array}
\right)^T
\end{equation}\\

The mapping function for the Rigid Deformation is given by

\begin{equation}
f_r(v) = |v-p_*|\frac{\overrightarrow{f_r}(v)}{|\overrightarrow{f_r}(v)|}+q_* 
\end{equation}
where $\overrightarrow{f_r}(v)=\sum_i \hat{q}_i A_i$. Because the rigid deformation was proposed to make the deformation be as rigid as possible, it performs best in our task. 

\subsubsection{Implementation Details of Shrinking}
\hspace{1.0em}
In detail, the control points $C_i(i=1\cdots 51)$ are exactly the landmarks for noses (15 points), mouths (13 points) and cheeks (21 points). In this paper, we achieve the global adjustment by controlling the moving vector $V_m^i$ for each control point. 

\begin{figure*}[t]
\begin{center}

    \includegraphics[width=0.7\linewidth]{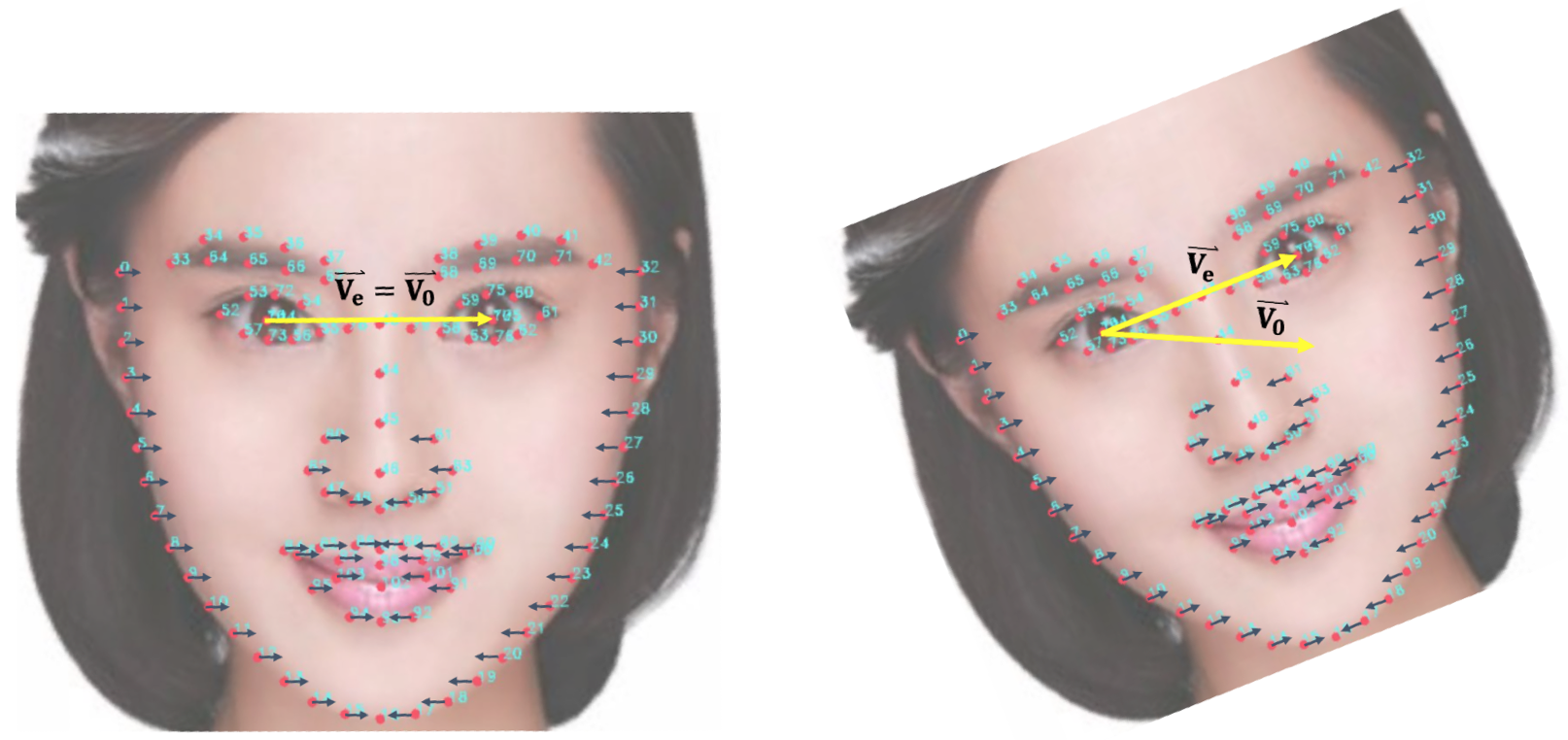}
\end{center}
   \caption{The shift of control points. Left: The face is horizontal. Right: The face is tilted and the moving vectors of the control points on the left side of the face have the same direction with $\protect\overrightarrow{V_e}$}
\label{shrink}
\end{figure*}

Before applying the mapping function, we need to make sure if the facial image is in the right direction. We decide this by checking if the vector $\overrightarrow{V_e}$ from the center landmark of the left eye $E_{left}$ to the center landmark of right eye $E_{right}$ is horizontal. To compare, we define the horizontal vector is $V_0$. The angle $\beta=<V_e,V_0>$ decides the direction of $V_m^i$. $V_m^i$ can be computed after knowing $\beta$ and the moving distance $l_i$. As shown in Figure \ref{shrink}, for control points on the left side of the face, $V_m^i=(l_i \cos\beta, l_i \sin\beta)$. For control points on the right side of the face, the situation is opposite and $V_m^i=(-l_i \cos\beta, -l_i \sin\beta)$. Keep the control points on the axes still which means $V_m^i=0$.


\section{Experimental Results}

The experiments mainly focus on showing the performance of different methods in the expansion and the shrinking. Two center point determination methods in the expansion and three MLS methods in the shrinking will be discussed in this section.

\subsection{Expansion Results}
\begin{figure*}[t]
\begin{center}

   \includegraphics[width=0.84\linewidth]{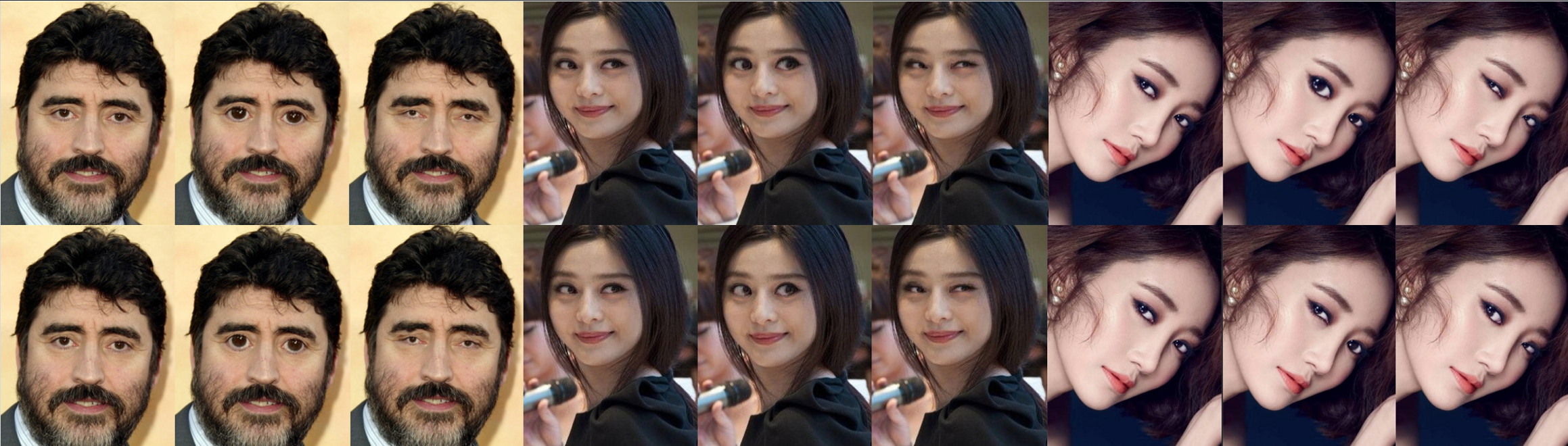}
\end{center}
   \caption{The results of the eye expansion process. Images in the first row are deformed images using the center eye landmark as the center point. Images in the second row are deformed images using the midpoint between $E_o$ and $E_d$ as the center point.}
\label{eye}
\end{figure*}

In general, the expansion operation is to make eyes bigger. Thus we set $a=50$ in this experiment to testify the expansion effect. However, we can also set $a$ to a negative value to make eyes smaller. We also regard this as a part of the expansion process and in this experiment, we set $a=-50$ for another set of results. We use the proposed two methods for center point determination. Three characteristic facial images are tested and the results are shown in Figure \ref{eye}. The three images from left to right depict people looking straightforward (normal), looking sideways (the pupil is not in the middle), and showing only the side face, in sequence. In the first row, the center point of the eye is the center eye landmark. In the second row, the center point is the midpoint between $E_o$ and $E_d$. For each set of images in each row, the first is the original image, the second is the deformed image when $a=50$, and the third is the deformed image when $a=-50$.

As shown in Figure \ref{eye}, the first set and the second set of images reveal nearly similar effects using different center point determination methods. However, two methods obtain different results when the input contains only the side face. The method using the midpoint of inner canthus and outer canthus as the center point is proposed to fight against the latent distortion that may be caused by the difference between the pupil and the real center point. But in some cases, the first method could also have a good performance as shown by the middle images of \ref{eye}, when the pupil is different from the real center point, the expansion effect using the first method is better than the effect of the second effect, which is beyond our expectation.

The reason for this result is that the line between the inner canthus and the outer canthus is often lower than the pupil. Thus, the center point is not accurate which makes the deformation also inaccurate. However, in practice, the second method can obtain better effects for some specific images. Thus, the choice of methods depends on the images when doing the expansion.

\subsection{Shrinking Results}
We compare results and the run-times of three methods: Affine Deformation, Similarity Deformation and Rigid Deformation in this subsection. We then evaluate how the weight $\alpha$ influences deformation results in Rigid MLS deformation.

\begin{figure}[htbp]
\begin{center}
   \includegraphics[width=0.8\linewidth]{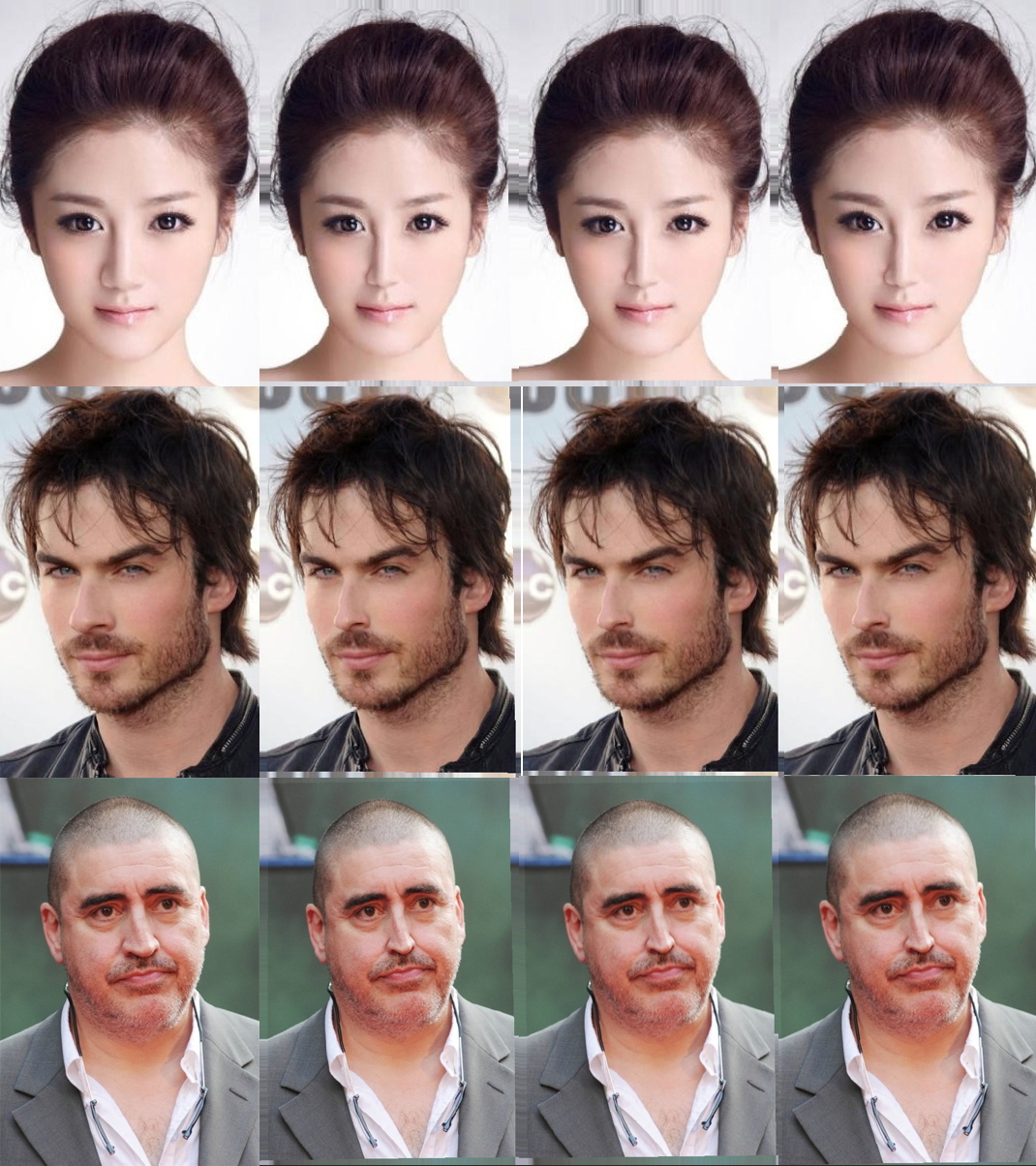}
\end{center}
   \caption{Comparison of three deformation methods. From left to right: origin, Affine MLS, Similarity MLS, Rigid MLS}
\label{result1}
\end{figure}

\paragraph{Deformation results comparison.}
As shown in Figure \ref{result1}, for images in each row, the first one is the original image, the second is the deformed image using Affine MLS, the third is the deformed image using Similarity MLS and the fourth is the deformation image using Rigid MLS. Affine MLS and Similarity MLS can both obtain quite satisfying results on the face, but inevitably induce other unstabilizing factors. For example, some distortion appears in the area close to the boundary of the deformed image. Rigid MLS can achieve a satisfying result which keeps the part we don't want to deform unchanged. However, because of the fact that these methods are all based on facial landmarks, a new problem is raised. Will results be affected if the face is not symmetric, especially when only the side face is revealed? One may think there will be distortion on the figures. However, when the figures are clear and the face does not turn a lot, the nose, the mouth and the cheeks can be shrunken successfully and little distortion can be perceived, and it can be proved by the second row in Figure \ref{result1}.

\paragraph{Run-time Comparison.}
Three methods have different run-times. As shown in Table \ref{table}, Similarity MLS and Rigid MLS's run-time is longer than Affine MLS's, but they are still edurable and welcome because they have better preformance on deformation results.

\begin{table}
\newcommand{\tabincell}[2]{\begin{tabular}{@{}#1@{}}#2\end{tabular}}
\begin{center}
\begin{tabular}{|l|c|c|c|}
\hline
Method & \tabincell{c}{Figure \ref{result1} \\ (top)} & \tabincell{c}{Figure \ref{result1} \\ (middle)}& \tabincell{c}{Figure \ref{result1} \\ (bottom)}\\
\hline\hline
Affine MLS& 0.49s & 0.53s & 0.64s\\
Similarity MLS& 0.88s & 0.89s & 1.06s\\
Rigid MLS& 0.90s & 0.89s & 1.06s\\
\hline
\end{tabular}
\end{center}
\caption{Deformation times for the various methods.}
\label{table}
\end{table}

\paragraph{Results with Different Weight.}

\begin{figure}[htbp]
\begin{center}

   \includegraphics[width=0.85\linewidth]{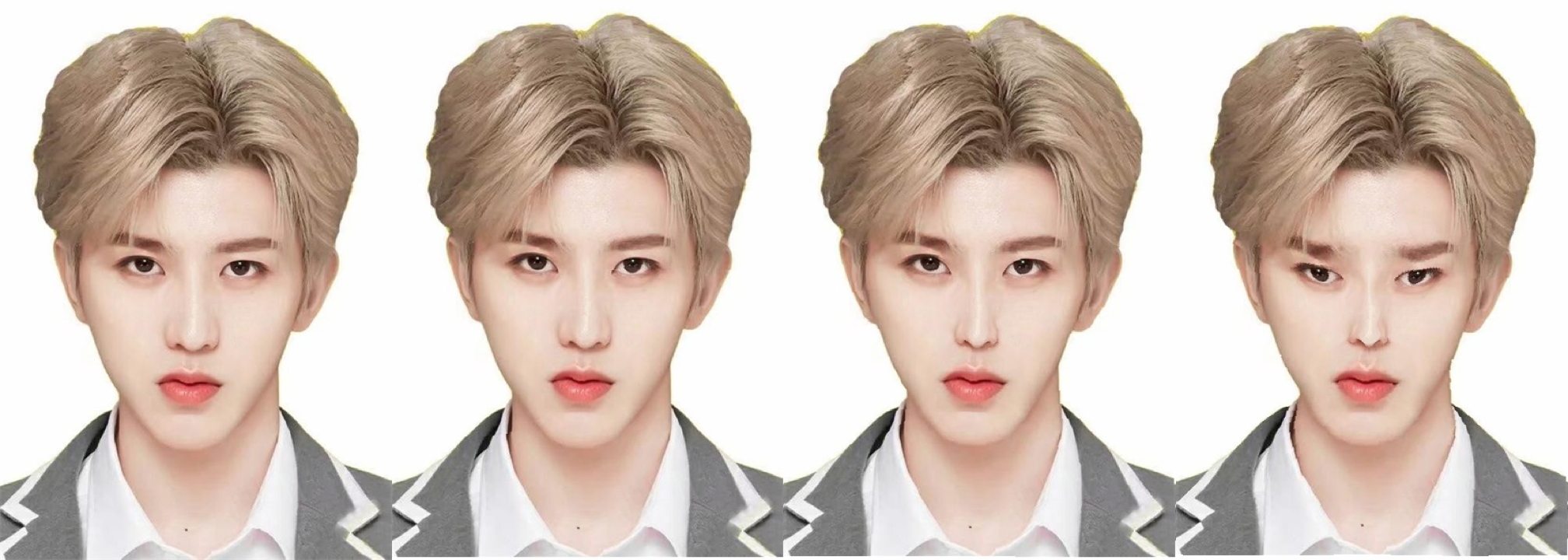}
\end{center}
   \caption{Comparison of rigid MLS with different weight. From left to right: origin, $\alpha=0.1$, $\alpha=1$, $\alpha=5$}
\label{result2}
\end{figure}

In this paper, we use the reciprocal of the distance from $v$ to the control point $q$ as the weight. The factor that affects the weight is $\alpha$ which be can known in Equation 6. To compare the effects of different $\alpha$, we try $\alpha=0.1,1,5$ for the same image. The different results are shown in Figure \ref{result2}.

As we can see, the deformed image is almost the same as the original image when $\alpha$ is small. However, the extent of deformation is too much when $\alpha$ is large. Therefore, we choose the most ideal situation when $\alpha=1$. Obviously, the deformation is exactly appropriate under such condition which can be shown by the third image in \ref{result2}.

\section{Conclusion}
In this paper, we describe a complete pipeline for facial image deformation based on facial landmark detection. The image deformation consists of two parts which are the expansion, based on bilinear interpolation and the shrinking, based on Moving Least Squares methods. During the experiments, we notice that the center points of eyes influence the expansion effect a lot. In addition, we compare the quality as well as the running time of deformed images using different MLS methods and then explore how the weight in Rigid Deformation influences the results of the shrinking process. Based on the experimental results, we find that the methods in both parts have a good performance. However, we think there is still substantial room for improvement. In future works we plan to explore more advanced methods and reduce the impact of inaccurate detection of facial landmarks.

{\small
\bibliographystyle{ieee}
\bibliography{paper}
}

\end{document}